\documentclass{ecai2014}
\usepackage{times}
\usepackage{latexsym}
\usepackage{graphicx}
\usepackage{color}
\usepackage{multirow}
\usepackage{amsmath}
\usepackage{amssymb}
\usepackage{subfig}
\usepackage[ruled,vlined,linesnumbered]{algorithm2e}

\newcommand {\IE} {\ensuremath {\mathbb{E}}}

\newcommand{\OPEN} {{\textsc{Open}}}

\newcommand{\lazyastar} {\ensuremath{LA^\ast}}
\newcommand{\rationallazyastar} {\ensuremath{RLA^\ast}}
\newcommand{\optcost} {\ensuremath{C^\ast}}

\SetKwBlock{Proc}{\{}{\}}

\newcommand{\comment}[1]{}

\title{Rational Deployment of Multiple Heuristics in IDA*}

\author{
David Tolpin \institute{CS Department, Ben-Gurion University. E-mail:shimony@cs.bgu.ac.il}
\and Oded Betzalel \institute{CS Department, Ben-Gurion University. E-mail:odedbetz@cs.bgu.ac.il}
\and Ariel Felner \institute{ISE Department, Ben-Gurion University. E-mail:felner@bgu.ac.il}
\and Solomon Eyal Shimony\institute{CS Department, Ben-Gurion University. E-mail:shimony@cs.bgu.ac.il}
}

\begin{document}

\maketitle

\begin{small}
\begin{abstract}
Recent advances in metareasoning for search has shown its usefulness in improving numerous search algorithms.
This paper applies rational metareasoning to IDA* when several admissible heuristics are available.
The obvious basic approach of taking the maximum of the heuristics is improved upon by lazy evaluation
of the heuristics, resulting in a variant known as Lazy IDA*. We
introduce a rational version of lazy IDA* that decides whether to compute the more expensive heuristics or to bypass it, based on a
myopic expected regret estimate.
Empirical evaluation in several domains supports the theoretical results, and
shows that rational lazy IDA*~is a state-of-the-art heuristic
combination method.

\end{abstract}
\end{small}


\section{Introduction}

Introducing  meta reasoning techniques into search is a research direction that has recently proved
useful for many search algorithms. All search algorithms have decision points on how to continue search.
Traditionally, tailored rules are hard-coded into the algorithms.
However, applying meta reasoning techniques based on value of information or other ideas can significantly speed up the search.
This was shown for depth-first search in CSPs~\cite{DBLP:conf/ijcai/TolpinS11},
for Monte-Carlo tree search~\cite{DBLP:conf/uai/HayRTS12}, and recently
for A* \cite{TOLPIN2013}. In this paper we apply meta reasoning techniques to speed up IDA* when
several admissible heuristics are available.

IDA*~\cite{BFID85} is a linear-space simulation of A*.
Thus it makes sense to examine how such speed-up was done for A* in a similar context,
as was done in Lazy A* (or \lazyastar, for short)~\cite{TOLPIN2013} by reducing the time spent on
computing heuristics.
A*  is a best-first heuristic search algorithm guided by the cost function $f(n)=g(n)+h(n)$.
A* uses OPEN and CLOSED lists and always expands the minimal cost node from OPEN, generates its children
and moves it to CLOSED.
When more than one admissible
heuristic is available, one can clearly evaluate all these heuristics, and use their {\em
maximum} as an admissible heuristic.
The problem with naive maximization is that all the heuristics are computed for all the generated nodes, resulting in
increased overhead, which can be overcome as follows.

With two (or more) admissible heuristics, when a node $n$ is
generated, Lazy A*  only computes one heuristic, $h_1(n)$, and adds $n$ to
\OPEN. Only when $n$ re-emerges as the top of \OPEN~is another heuristic, $h_2(n)$, evaluated;
and if $h_2(n)>h_1(n)$ then $n$ is re-inserted into \OPEN.
If the goal is reached before node $n$'s re-emergence, the computation
of $h_2(n)$ is never performed, thereby saving time, especially if $h_2$ is computationally heavy.
In {\em rational lazy} A*~(\rationallazyastar)~\cite{TOLPIN2013}, the ideas of lazy heuristic
evaluation and trading additional node expansions for decreased time for computing  heuristics were combined.
\rationallazyastar~is based on rational meta-reasoning, and uses a myopic {\em
regret} criterion to decide whether to compute $h_2(n)$ or to
bypass the computation of $h_2$ and expand $n$ immediately when $n$ re-emerges
from \OPEN. \rationallazyastar~aims at reduced search time, even at the expense of more node expansions than \lazyastar.

The memory consumption of A* is linear in the number of generated nodes, which is typically exponential in
the problem description size, which may be unacceptable.
In contrast to A*, IDA* is a linear-space algorithm which emulates A* by performing a series of depth-first searches from the root,
each with increasing costs, thus re-expanding nodes multiple times.
IDA* is typically used in domains and problem instances where A*~requires more than the available memory and thus cannot
be run to completion. If the heuristic $h(n)$ is admissible (never overestimates the real cost to the goal)
then the set of nodes expanded by A*~is both necessary and sufficient to find the optimal path to the goal~\cite{ASTR85}.
Similar guarantees holds for IDA*~ under some additional reasonable assumptions.
Thus, techniques used to  develop \rationallazyastar, can in principle be applied to IDA*, the focus of this paper.
However, IDA* has a different logical structure and needs a completely different treatment.
In particular, in A* one needs to assign an $f$-value to each generated node $n$ while in IDA*, one only needs to know
whether the $f$-value is below or above the current threshold.

The first thing to consider for IDA* is lazy evaluation of the heuristics.
In order to reduce the time spent on heuristic computations, Lazy IDA*~evaluates
the heuristics one at a time, lazily. When $h_1$ causes a cutoff there is no need to evaluate $h_2$.
Unlike Lazy A*, where lazy evaluation must pay an overhead (re-inserting into the OPEN list)
\cite{TOLPIN2013},
Lazy IDA* (LIDA*) is straightforward and has no immediate overhead.

The main contribution of this paper is {\em Rational lazy IDA*} (RLIDA*) which
uses meta reasoning at runtime\footnote{This paper is an extended version of a short (2-page) paper
to appear in the ECAI 2014 proceedings. In addition to containing all the analysis that could not fit into the short version,
there are some additional experimental results and a comparison to additional related work.}.
We analyze IDA* and provide a
criterion, based on a myopic expected regret, which decides whether to evaluate a heuristic or to bypass
that evaluation and expand the node right away.
We then provide experimental results on sliding tile puzzles and on the container relocation problem \cite{Zhang:2010:IIA:1945758.1945763},
showing that RLIDA* outperforms both IDA* and LIDA*.

\section{Lazy IDA*}

We begin by describing IDA*, and the minor change needed to make it use the heuristics
lazily, thus implementing lazy IDA*.

\subsection{Definitions}

Throughout this paper we assume for clarity that we have two available admissible heuristics, $h_1$ and $h_2$.
Unless stated otherwise, we assume that $h_1$ is faster to compute than $h_2$
but that $h_2$ is {\em weakly more informed}, i.e., $h_1(n) \leq h_2(n)$ for
the majority of the nodes $n$, although counter cases where $h_1(n) > h_2(n)$
are possible.
We say that $h_2$ {\em dominates} $h_1$, if such counter cases do not
exist and $h_2(n) \geq h_1(n)$ for {\em all} nodes $n$.
We use $f_1(n)$ to denote $g(n)+h_1(n)$, and $f_2(n)$
to denote $g(n)+h_2(n)$.
We denote the cost of the optimal solution by $\optcost$. Additionally, we
denote the computation time of $h_1$ and of $h_2$ by $t_1$ and $t_2$, respectively.
Unless stated otherwise we assume that $t_2$ is much greater
than $t_1$. We thus mainly aim to reduce the number of times $h_2$ is computed.

\subsection{Why use lazy IDA*?}

Let $T$ be the IDA* threshold. After $h(n)$ is evaluated, if $f(n)=g(n)+h(n)> T$, then $n$ is pruned and IDA*
backtracks to $n$'s parent. Given both $h_1$ and $h_2$, a naive implementation of IDA* will evaluate them both and
use their maximum in comparing against $T$.
Lazy IDA* (LIDA*) is based on the simple fact that when you have an {\em or} condition in
the form of $cond1~or~cond2$ then if $cond1 = True$ then $cond2$ becomes irrelevant (don't-care) and need not be computed,
as the entire {\em or} condition is surely true.
In the context of IDA*, if $f_1(n) > T$ then the search can backtrack without the need to compute $h_2$.
This simple observation is probably recognized by most implementers of IDA*.
Thus, it is likely that LIDA* is a way to implement
IDA* when more than one heuristic is present.

\begin{algorithm}[t]
Lazy-IDA* (root) \Proc{
   Let Thresh = max($h_1$(root), $h_2$(root)) \\
   Let solution = null\\
   \While{solution $==$ null and Thresh $< \infty$}{
       solution, Thresh = Lazy-DFS(root, Thresh)
   }
   \Return solution
}

Lazy-DFS(n, Thresh) \Proc{
    \If{g(n) $>$ Thresh}{
      \Return null, g(n)}
    \If{goal-test(n)}{
      \Return n, Thresh}
    \If{g(n)+$h_1$(n) $>$ Thresh}{
      \Return null, g(n)+$h_1$(n)}

    \If{{\bf opt-cond} and g(n)+$h_2$(n) $>$ Thresh}{\label{RLIDA:opt-cond}
      \Return null, g(n)+$h_2$(n)}

    Let next-Thresh $= \infty$\\
    \For{n' in successors(n)}{
      Let  solution, temp-Thresh = Lazy-DFS-lim(n', Thresh)\\
      \If{solution   $\neg =$ null} {
        \Return {solution, temp-Thresh}}
      \Else{ Let next-Thresh = min(temp-Thresh, next-Thresh)}
     }
     \Return{null, next-Thresh}
}

\caption{Lazy IDA*}
\label{RLIDA}
\end{algorithm}

The pseudo-code for  LIDA*~is depicted as Algorithm \ref{RLIDA}. In lines 13-14 we check whether $f_1$ is already
above the threshold in which case, the search backtracks. $h_2$ is only calculated (in lines 15-16)
if $f_1(n) \leq T$. The ``optional condition'' in line~\ref{RLIDA:opt-cond} is needed for the rational lazy
A*~algorithm, described below, which entails adding
appropriate conditions that aim at $h_2$ only if its usefulness
outweights its computational overhead on average. In the standard version of lazy
IDA*, the ``optional condition'' in line~\ref{RLIDA:opt-cond} is always true, and the respective
heuristics are always evaluated at this juncture. We also note that lines 9-10
are needed to ensure that the goal test at lines 11-12 will only return the optimal solution.
This check is particulary needed for Rational Lazy IDA* as described below.

\subsection{Issues in Lazy IDA*}

Several additional obvious improvements to LIDA* are possible. Here we examine some such
potential enhancements, as well as possible pitfalls.

\subsubsection{Heuristic bypassing}

{\em Heuristic bypassing} (HBP) is a technique that in many cases allows
bypassing the computation of a given heuristic without causing any other change in the course of the algorithm.
In A* one needs to compute an $f$-value, while 
Applied to IDA*, one only needs to know whether the $f$-value is below
or above the threshold. First, it is important to note that Lazy IDA* as described above,
is a special case of HBP. When $f_1(n) > T$ there is no need to consult $h_2(n)$ and we bypass
the computation of $h_2$. Another variant of HBP for LIDA* is applicable
for a node $n$ under the following two preconditions: {\bf (1)} the
operator between $n$ and its parent $p$ is bidirectional, and {\bf (2)} both
heuristics are {\em consistent}~\cite{INCJUR}.
Suppose that node $n$ was generated and that $p$ is the parent of $n$;
that the cost of the edge is $c$ and that $f_1(p) + c < f_2(p)$.  Since $p$ was expanded, we know
that $f_2(p) \leq T$. Since the heuristics are consistent,
we know that $f_1(n) \leq f_1(p) + c \leq T$. Thus, in such cases, one can skip the computation of $h_1(n)$ and go directly to $h_2$.
Nevertheless, the savings here are negligible as we assumed that $t_1 << t_2$ and our aim is thus to decrease
the number of times $h_2$ is computed. We also note that HBP needs additional effort for book keeping.

When the heuristic is inconsistent
then a mechanism called bidirectional pathmax (BPMX) can be used to propagate heuristic
values from parents to children and vice versa~\cite{INCJUR}.
Using exhaustive evaluations of all heuristics, even if $h_1(n)$ already exceeded the threshold,
can potentially help in propagating larger heuristic values to the neighborhood of $n$.
Nevertheless, experiments showed that even in this context, lazy evaluation of heuristics
is faster in time than exhaustive evaluation~\cite{INCJUR}.

\subsubsection{Extra iterations of Lazy IDA*}

In rare cases, LIDA* can cause extra DFS iterations. Suppose that the current threshold is $T$ and
the current value of the {\em next threshold} (NT) is $T+3$ as some node $m$ was seen in the current
iteration with $f(m) = T+3$. Now we generate node $n$ with $f_1(n)= T+1$ and thus set $NT=T+1$ and bypass $h_2$.
However, if $f_2(n)=T+2$ then consulting $h_2$ would have caused $NT=2$. With LIDA*, we may now start a new and redundant
DFS iteration with $T+1$.

While Lazy A*, was always as informative as A* using the maximum of the heuristics,
this is not the case for Lazy IDA*.
Nevertheless, since there is potentially an exponential number of nodes in the frontier of a DFS iteration,
such scenarios are quite rare and Lazy IDA* outperforms regular IDA* despite this worst-case scenario.

\section{Rational Lazy IDA*}

A general theory for applying rational meta-reasoning for search algorithms was presented in~\cite{RussellWefald}.
Using principles of rational meta-reasoning theoretically every algorithm action (heuristic function evaluation, node
expansion, open list operation) should be treated as an action in a sequential
decision-making meta-level problem: actions should be chosen so as to
achieve the minimal expected search time. However, the appropriate
general meta-reasoning problem is extremely hard to define precisely and
to solve optimally. In order to apply it practically, specific assumptions and simplifications should be added.

In this paper we focus on just one decision type,
made in the context of IDA* - that of deciding whether to evaluate or to bypass the computation of $h_2$.
In order to choose rationally, we define a criterion based on the regret for bypassing
$h_2(n)$ in this context. We define regret here as the value lost (in terms of increased run time)
due bypassing the computation of $h_2(n)$, i.e. how much runtime is increased due to bypassing the computation.
We wish to compute $h_2(n)$ only if this regret is positive on the average.
Some of the ideas behind Rational Lazy IDA* are borrowed from those of
\cite{TOLPIN2013} and Rational Lazy A* (RLA*). However, the assumptions of RLA* are different,
and cannot be used for IDA* as they were made under the assumption that there exists an
OPEN list and that an $f$-value of a node
should be stored within the node. In contrast, in IDA* there is no OPEN list and we only need to know whether $f(n)$ is below
or above the threshold $T$. Therefore IDA* needs a different treatment.

In IDA*, each iteration is a depth-first search up to a
gradually increasing threshold $T$, until a solution is found. For each node $n$, we say that evaluating $h(n)$ is
{\em helpful} if $g(n)+h(n)>T$. That is, the heuristic {\em helped} in the
sense that node $n$ is pruned, rather than expanded, in this iteration.

The only addition of Rational Lazy IDA* to Lazy IDA* is the option to bypass $h_2(n)$ computations~(line~\ref{RLIDA:opt-cond}).
In this case, $n$ is expanded right away.\footnote{It is important to note that in
such cases, $f_2(n)$ might be greater than $T$.
For this reason we added lines 9-10 in the pseudo code
above, to ensure that the solution returned is always optimal.}
Suppose that we choose to compute $h_2$ --- this results in one of the following outcomes:
\begin{enumerate}
\item $h_2(n)$ is not helpful and $n$ is immediately expanded.
\item $h_2(n)$ is helpful (because $g(n)+h_2(n) > T$), pruning $n$, which is not expanded in the current IDA* iteration.
\end{enumerate}

Observe that computing $h_2$ can be {\em beneficial} only in outcome 2 plus the additional
condition that the time saved due to pruning a search subtree outweighs the time to compute $h_2$, i.e., $t_2(n)$. However, whether outcome 2 
takes place after a given state is not known to the algorithm until {\em after} $h_2$ is computed. The algorithm must decide whether to evaluate $h_2$
according to what it \textit{believes to be} the probability of each
of the outcomes. The time wasted by being sub-optimal in deciding
whether to evaluate $h_2$ is called the {\em regret} of the decision.
We derive a \textit{rational policy} for deciding when to evaluate
$h_2$, under the following assumptions:

\begin{enumerate}
\item The decision is made \textit{myopically}: we work under the belief that the algorithm continues to  behave like Lazy IDA* starting with the children of $n$.
\item $h_2$ is \textit{consistent}: if evaluating $h_2$ is beneficial
  on $n$, it is also beneficial on any successor of $n$.
\item As a first approximation, we also assume that $h_1$ will not cause pruning in any of the children.
\end{enumerate}

If Rational Lazy IDA* is indeed better than Lazy IDA*, the
first assumption results in an upper bound on the regret. Note that these meta-reasoning assumptions are made in order to derive
decisions, and as is common in research on meta-reasoning, the assumptions do {\em not} actually hold in practice~\cite{RussellWefald}.
Nevertheless, if the violation of the assumptions is not ``too severe'', the resulting algorithms
still show significant improvement. Without such assumptions the model becomes far too complicated and one cannot move ahead at all.
For example, the myopic assumption trivially does not hold by design, as applying it strictly at runtime means that we only
use the rational decision rule at the root, which does not make sense in practice. Violating this assumption results in an actual expected runtime
that is lower than that computed under this assumption. The other two simplifying assumptions do not have this nice property as far as we know, however,
and one would prefer to drop them. This non-trivial issue remains for future research.

If $h_2(n)$ is not helpful and we decide to compute it, the effort invested in evaluating $h_2(n)$ turns out to be wasted. On the other hand, if $h_2(n)$ is
helpful but we decide to bypass it, we needlessly expand $n$. Due to
the myopic and other assumptions, Rational Lazy IDA* would evaluate both $h_1$
and $h_2$ for all children of $n$. 
Due to consistency of $h_2$, the children of $n$ will not be expanded in this IDA* iteration.
\begin{table}
\begin{center}
\begin{tabular}{|l|c|c|}
\hline
               & Compute $h_2$ & Bypass $h_2$\\
\hline
$h_2$ helpful &   0            & $t_e+b(n)t_1+(b(n)-1)t_2$\\
\hline
$h_2$ not helpful & $t_2$      & 0 \\
\hline
\end{tabular}
\end{center}
\caption{Time losses in Rational Lazy IDA*}
\label{tbl:rlida-rational-lazy-ida-time}
\end{table}

Table~\ref{tbl:rlida-rational-lazy-ida-time}
summarizes the regret of each possible decision, for each possible future
outcome; each column in the table represents a decision, while each row
represents a future outcome.
In the table, $t_e$ is the time to  evaluate $h_1$  and expand $n$, and b(n) is the local branching factor at node n (taking into account parent pruning). Computing $h_2$ needlessly ``wastes'' $t_2$ time.
Bypassing $h_2$ computation when $h_2$ would have been helpful ``wastes''
$t_e+b(n)t_1+b(n)t_2$ time, but because computing $h_2$ would have cost $t_2$, the
regret is $t_e+b(n)t_1+(b(n)-1)t_2$.

Let us denote the probability that $h_2(n)$ is helpful by
$p_h$. The expected regret of computing $h_2(n)$ is thus $(1-p_h) t_2$.
On other hand, the expected regret of bypassing $h_2(n)$ is $p_h(t_e+b(n)t_1+(b(n)-1)t_2)$.
As we wish to minimize the expected regret, we should thus evaluate $h_2$ just when:
\begin{equation}
(1-p_h) t_2 < p_h (t_e+b(n)t_1+(b(n)-1)t_2)
\end{equation}
or equivalently:
\begin{equation}
(1- p_h b(n)) t_2 < p_h (t_e + b(n)t_1)
\label{eq:simplified}
\end{equation}

If $p_h b(n) \ge 1$ (the left side of the equations is negative), then the expected regret is minimized by always
evaluating $h_2$, regardless of the values of $t_1$, $t_2$ and $t_e$. A simple decision rule would be
to evaluate $h_2$ exactly in these cases.

For $p_h b(n) < 1$,  the decision of whether to evaluate $h_2$
depends on the values of $t_1$, $t_2$ and $t_e$:
\begin{equation}
\mbox{\bf evaluate }h_2\mbox{ \bf if }t_2<\frac {p_h} {1-p_hb(n)} (t_e+ b(n)t_1)
\label{eqn:rlida-criterion-general}
\end{equation}
The factor $\frac {p_h} {1-p_hb(n)}$ depends on the potentially unknown
probability $p_h$, making it difficult to reach the optimum decision.
However, if our goal is just to do better than Lazy IDA*, then it is
safe to replace $p_h$ by an upper bound on $p_h$. We discuss this next.

\subsection{Bounding the probability that $h_2$ is helpful}\label{sec:bounds}

Search time can be saved by evaluating $h_2$ selectively, only in the
nodes where the probability that the evaluation is helpful is
``high enough''. 
In particular, in the case of two heuristics, $h_1$ and
$h_2$, the decision whether to evaluate $h_2(n)$ can be made
based on $h_1(n)$ and prior history of evaluations of $h_1$ and $h_2$
on the same or ``similar'' nodes.
One can try to estimate $p_h$, either online or offline in order
to use the decision boundaries such as Equation \ref{eqn:rlida-criterion-general}
based on these empirical frequencies directly.

Nevertheless, we examine another possibility here,
based on the rationale that our goal in RLIDA* is to do better than simple LIDA*,
and wish to trade off computation times ``safely'', i.e. with little risk of being
{\em worse} than LIDA*.
One way to estimate the probability $p_h$ that the evaluation
is helpful ``safely'' is to bound this probability using concentration
inequalities. 

Concentration inequalities bound probabilities of certain events for
a bound random variable, that is, such a variable $x$ for which
$\Pr[x\in [0,1]] = 1$, and we need to construct such a variable. Let
$x$ be:
\begin{equation}
x=1-\frac {h_1(s)} {\max(h_1(s),h_2(s))}
\label{eq:x}
\end{equation}
It is easy to see that $x\in[0,1]$ and increases with $h_2(s)$. The
condition $h_2(n)>T-g(n)$ (i.e., $h_2(n)$ is helpful) is equivalent to condition $x>l$ where:
\begin{equation}
l=1-\frac {h_1(n)} {T-g(n)}
\label{eq:l}
\end{equation}
We need to bound the probability that $\Pr(X_{N+1}>l)$ given the prior
history of evaluations of $x$ (that is, of $h_1$ and $h_2$). Denote by
$\overline{x_N}$ the average of $N$ samples:
\begin{equation}
\overline{x_N}=\frac 1 N \sum_{i=1}^N X_i
\label{eq:sample-average}
\end{equation}
The probability $\Pr(X_{N+1}>l)$ is less than the probability that the
mean $\IE[x]$ of the random variable $x$ is at least
$\mu,\,\overline {x_N}\le\mu\le l$, plus the probability that $X_{Nn+1}>l$ given
$\IE[x]=\mu$ (the union bound).
\begin{equation}
p_h=\Pr(X_{N+1}>l) \le \Pr(\IE[x]> \mu) + \Pr(X_{N+1}>l \,\vline\, \IE[x]=\mu)
\label{eq:x-gt-l}
\end{equation}
Denote $\mu=(1-\alpha)\overline{x_N}+\alpha l,\,\alpha \in [0,1]$ ---
we will obtain the bound as a function of $\alpha$ and then select
$\alpha$ that minimizes the bound. According to the Hoeffding inequality:
\begin{equation}
\Pr(\IE[x] > (1-\alpha)\overline {x_N}+\alpha l)<e^{-2 N (\alpha (l-\overline {x_N}))^2}
\label{eq:hoeffding}
\end{equation}
and to the Markov inequality:
\begin{equation}
\Pr(X_{N+1}>l\,\vline\,\IE[x]=(1-\alpha)\overline{x_N}+\alpha l) < \frac
   {(1-\alpha)\overline {x_N}+\alpha l} l
\label{eq:markov}
\end{equation}

An upper bound for the probability $\Pr(X_{N+1}>l)$ is a function of $\alpha$:

\begin{equation}
p_h=\Pr(X_{N+1}>l)\le B(\alpha)=e^{-2 N (\alpha (l-\overline  {x_N}))^2}+\frac {(1-\alpha)\overline {x_N}+\alpha l} l
\label{eq:pr-star}
\end{equation}

The bound $B(\alpha)$ can be minimized for $\alpha\in[0,1]$ by solving
$\frac {dB(\alpha)} {d\alpha}=0$, but a closed-form
solution does not generally exist. However, a reasonable
value for $\alpha$ can be easily found. Choosing
\begin{equation}
\alpha^*=\frac {\sqrt {\frac {\log \sqrt{2N} l} {2N}}} {l-\overline {x_N}}
\label{ex:alpha}
\end{equation}
and substituting into (\ref{eq:pr-star}), obtain
\begin{equation}
\Pr(X_{N+1}>l) \le B^*=B(\alpha^*)=\frac {1+\sqrt{\log{\sqrt{2N}l}}} {\sqrt {2N} l} +
\frac {\overline {x_N}} l
\label{ex:p-bound}
\end{equation}
In the bound (\ref{ex:p-bound}) the second term
$B_2=\frac {\overline  {x_N}} l$ is tantamount to the Markov
inequality when the sample average $\overline{x_N}$ coincides with the mean
$\IE[x]$. The first term $B_1=\frac {1+\sqrt{\log{\sqrt{2N}l}}} {\sqrt  {2N} l}$
does not depend on $x$ and for $l>0$ approaches zero
as $N$ approaches infinity. Although the
concentration inequalities are correct for iid samples, a state that does not
necessarily hold for samples of heuristic values during the search, nevertheless
it is a usable first-order approximation. We use $B^*$ as defined in Equation \ref{ex:p-bound}
as an estimate of $p_h$.






\section{Empirical evaluation}\label{sec:empirical}

The greatest advantage of IDA* over A* is storage complexity.
However, IDA* has a number of limitations. First, the number of
nodes expanded by IDA* is typically much greater than that of A* because IDA* is
unable to detect transpositions and because in every iteration, IDA* repeats the former
iterations. In addition, IDA* preforms very poorly if there is a large
number of different $f$-costs below $C^*$ encountered during the search
(leading to a large number of iterations), which occurs in domains such as TSP.


Therefore we selected for empirical evaluation domains that are known to be IDA*-friendly (such as the 15-puzzle),
or where recent work has shown IDA* to perform well, such as the container relocation
problem \cite{Zhang:2010:IIA:1945758.1945763}. Regretfully, most planning problems (from the planning competitions)
used in~\cite{TOLPIN2013}, are inappropriate for IDA* due to multiple transpositions in the search space. 
Another requirement we had is the availability
of known informative admissible heuristics for the domain (otherwise it does not pay to compute them),
that are costly to compute (if they are very cheap, we might as well always compute them). In domains
where the latter requirements do not hold, elaborate meta-reasoning on whether to compute a heuristic will thus
obviously not achieve any significant improvement.

The above restrictions are obvious limitations to the applicability of the scheme proposed in this paper, and
should be considered when trying to apply our methods. Nevertheless, as stated in Section \ref{sec:related},
our scheme should be extensible to other IDA*-like algorithms where the large number of f-costs is
not a problem.

\subsection{Sliding tile puzzles}

We first provide evaluations on the 15-puzzle and its weighted variant,
where the cost of moving each tile is equal to the number on the tile.
Note that there is another version of the weighted 15-puzzle where the cost of a tile move
is the {\em reciprocal} of the number on the tile~\cite{thayer:bss}. However, the number of possible f-costs
under f* in this version is typically very large, thus in this reciprocal variant, IDA* is expected to
perform abysmally, making IDA* inapplicable to this domain. Indeed, some preliminary runs confirmed this expectation, and
we therefore dropped this reciprocal weights version from our evaluation.

For consistency of comparison, we used as test cases for the 15 puzzle
98 out of Korf's 100 tests~\cite{BFID85}: all the tests that
were solved in less than 20 minutes with standard IDA*  using the
Manhattan Distance (MD) heursitic. (All experiments were performed using Java, on a 3.3GHz
AMD Phenom II X6 1100T Processor, with 64 bit Ubuntu 12.04, and with sufficient memory to avoid paging.)
As the more informative heuristic $h_2$  we used the {\em linear-conflict heuristic} (LC)~\cite{KorfT96}
which adds a value of 2 to MD for pairs of tiles that are in the same row (or the same column)
as their respective goals but in a reversed order. One of these tiles will need to move away
from the row (or column) to let the other pass.

\begin{table}
\begin{centering}
\begin{tabular}{|l | r | r | r | r| }
\hline
algorithm & time & generated & $h_2$ total & $h_2$ helpful \\
 \hline
IDA* (MD)            & 58.84  & 268,163,969  &  &  \\
IDA* (LC)            & 40.08 & 30,185,881  &  &  \\
LIDA*                & 32.85 & 30,185,881  & 21,886,093 & 6,561,972 \\
RLIDA*               &  20.09 & 47,783,019 &  8,106,832 & 4,413,050 \\
Clairvoyant          &  12.66 &  30,185,881 & 6,561,972 & 6,561,972 \\
\hline
\end{tabular}
\caption{15 puzzle}
\label{tbl:15-puzzle-results}
\end{centering}
\end{table}

Since the runtime of both heuristics is nearly constant across the states, (i.e., $t_1(n) \approx c_1$ and $t_2(n)  \approx c_2$
for some constants $c_1, c_2$)
it turns out that the decision of whether to compute $h_2$ is stable across a wide range of $p_h$ values, and thus a constant value
of $p_h$ performs well for this domain.
Results are presented for an assumed constant $p_h=0.3$, estimated offline from trial runs of RLIDA* on a few problem instances.
Average results for IDA* with only MD, IDA* with LC, Lazy IDA* using both heuristics, and Rational Lazy IDA*, are
shown in Table~\ref{tbl:15-puzzle-results}. The advantage of Rational Lazy IDA* is evident: even though it expands
many more nodes than Lazy IDA*, its runtime is significantly lower as it saves even more time on evaluations of LC. LIDA*
evaluated LC 21,886,093 times, out of which only 6,561,972 were helpful.
Much time was wasted on evaluating non-helpful heuristics. In contrast,
RLIDA* only chose to evaluate LC 8,106,832 times, out of which 4,413,050 were helpful.
The bottom {\em Clairvoyant} row is an unrealizable scheme that uses an oracle,
not achivable in practice, which has a runtime better than any achievable optimal decision on whether to evaluate $h_2$.
Its numbers were estimated by using the LIDA* results, assuming that $h_2$ was computed only in the 6,561,972 helpful
nodes, and bypassed otherwise. 
As can be seen, the runtime of our version of RLIDA* is closer to Clairvoyant than to LIDA*.
It shows that much of the potential of RLIDA* was indeed exploited by our version.

\begin{table}
\begin{centering}
\begin{tabular}{|l | r | r | r | r| }
\hline
algorithm & time & generated & $h_2$ total & $h_2$ helpful \\
 \hline
IDA* (MD)               & 184.46 & 822,898,188 &  &  \\
IDA* (LC)               & 155.35 & 104,943,867 &  &  \\
LIDA*                   & 112.74 & 104,943,890 & 65,660,207 & 12,549,104 \\
RLIDA*                  & 63.08 &  137,881,842 & 21,564,188 & 8,871,727 \\
Clairvoyant             & 40.36 &  104,943,890 & 12,549,104 & 12,549,104  \\
\hline
\end{tabular}
\caption{Weighted 15 puzzle}
\label{tbl:weighted-15-puzzle-results}
\end{centering}
\end{table}

Table \ref{tbl:weighted-15-puzzle-results} shows similar results for 82 of the previous initial positions on weighted 15 puzzle that
were solved in 20 minutes by IDA* (the weighted 15 puzzle is harder).
In this domain, Rational Lazy A* also achieves a significant speedup and was much closer to Clairvoyant than to LIDA*.

For the heuristics we used in our tests and for $p_h=0.3$, it turns out that the decision on whether to evaluate $h_2$ depends
just on the branching factor: evaluate $h_2$ only for $b(n)=3$ (excluding the  parent), i.e. for cases where the blank was in the middle.
Applying the bounds from Section \ref{sec:bounds}
to estimate $p_h$ did not achieve significant further improvement over RLIDA* with a constant $p_h$ (not shown in the tables), due to the fact that
the simple decision rule was rather stable across a relatively wide range of $p_h$. We thus expect this same rule to work
for sliding tile puzzles of other dimensions, and tried the same scheme in rectangular tile puzzles: 3*5
(numbers from 1 to 14) and 3*6 (numbers from 1 to 17). Since the fraction of nodes with 3 children in these puzzles is lower than the
4*4 puzzle, we expect RLIDA* to do better than in the 4*4 puzzle.
As we did not have access to standard benchmark instances,
we generated instances using
random walks of 45 to 80 steps from the goal state. 
\begin{table}
\begin{centering}
\begin{tabular}{|l | r | r | r | r| }
\hline
algorithm & time & generated & $h_2$ total & $h_2$ helpful \\
\hline
IDA* (MD) & 134.27 & 518,625,911 & & \\
IDA* (LC) & 68.65 & 53,073,488 & & \\
LIDA* & 59.89 & 53,073,499 & 36,000,253 & 8,218,490 \\
RLIDA* & 38.31 & 77,199,730 & 12,104,449 & 6,564,049 \\
Clairvoyant & 27.99 & 53,073,499 & 8,218,490 & 8,218,490 \\
\hline
\end{tabular}
\caption{3 by 5 puzzle}
\label{tbl:3-On-5-puzzle-results}
\end{centering}
\end{table}
\begin{table}
\begin{centering}
\begin{tabular}{|l | r | r | r | r| }
\hline
algorithm & time & generated & $h_2$ total & $h_2$ helpful \\
\hline
IDA* (MD) & 17.76 & 66,655,434 & & \\
IDA* (LC) & 30.11 & 17,098,738 & & \\
LIDA* & 21.99 & 17,098,746 & 10,308,664 & 1,473,548 \\
RLIDA* & 10.68 & 21,053,303 & 2,882,141 & 1,007,129 \\
Clairvoyant & 7.17 & 17,098,746 & 1,473,548 & 1,473,548 \\
\hline
\end{tabular}
\caption{3 by 6 puzzle}
\label{tbl:3-On-6-puzzle-results}
\end{centering}
\end{table}

Tables \ref{tbl:3-On-5-puzzle-results}, \ref{tbl:3-On-6-puzzle-results} show that the improvement factor in both domains
due to rational lazy IDA* is similar to that obtained in the (4*4) 15 puzzle. However the gap between RLIDA* and the unrealizable
clairvoyant scheme is smaller than for the 4*4 puzzle, so RLIDA* seems to be making better decisions in these latter variants, as expected.
Though indicative, one caveat is that the way instances were generated in the rectangular versions is different from the 4*4 puzzle,
and the general shape of the search space may also differ.

\subsection{Container relocation problem}

The container relocation problem is an abstraction of a planning problem encountered in retrieving stacked containers
for loading onto a ship in sea-ports \cite{Zhang:2010:IIA:1945758.1945763}.
We are given $S$ stacks of containers, where each stack consists of up to $T$ containers.
In each stack, containers are stacked on top of one another.
In the initial state there are $N \leq S \times T$ containers, arbitrarily numbered from 1 to $N$.
The rules of stacking and of moving containers is the same as for blocks in the well-known blocks world domain, i.e., a
container can be moved if there is no container on top of it. However, unlike blocks-world planning, the objective function
is different, as follows.

The goal is to retrieve all containers in order of number, from 1 to $N$, where ``retrieve''
can be seen as placing a container on an additional, special and always empty, stack where the container disappears
(in the application domain this ``special stack'' is actually a freight truck that takes the container away
to be loaded onto a ship). The objective function to minimize is the number of container moves until all containers are gone (``loaded
onto the truck''). The complication comes from the fact that we can only ``retrieve'' a container if it is at the top of one of
the stacks. Thus, containers on top of it should be moved away.
Optimally solving this problem is NP-hard \cite{Zhang:2010:IIA:1945758.1945763}.

Although there are various variants of this problem, we assume here the version where each
container (``block'' in blocks-world terminology) is uniquely numbered. Another assumption typically made
is that a stack $s$ that currently has $T$ containers is ``full'' and no additional containers can
be placed on $s$ until some container is moved away from $s$. We also address only the ``restricted'' version
of the problem \cite{Zhang:2010:IIA:1945758.1945763}, where the only relocations allowed are of containers currently on top of
the smallest numbered container.
Finally, since a solution always involves
removing all $N$ containers, and each container can be moved to the truck only once, it is customary
to count only moves from stack to stack (called ``relocations''), ignoring the final move of
containers to the truck.

The heuristics we used for the experiments are as follows. Every container numbered $X$ which is above
at least one container $Y$ with with a number smaller than $X$
must be moved from its stack in order to allow $Y$ to be retrieved. The number of such containers
in a state can be computed quickly, and forms an admissible heuristic
denoted $LB_1$ in \cite{Zhang:2010:IIA:1945758.1945763}. A more complicated heuristic adds one relocation for
each container that must be relocated
a second time as any place to which it is moved will block some other container.
Following \cite{Zhang:2010:IIA:1945758.1945763}, we denote this heuristic
by $LB_3$\footnote{To guarantee admissibility we made some minor
notation changes from how this heuristic is formally stated in the original paper}.
This heuristic requires much more computation time than $LB_1$, and additionally its runtime depends heavily on the state.

\begin{table}
\begin{centering}
\begin{tabular}{|l | r | r | r | r| }
\hline
algorithm & time & generated & $h_2$ total & $h_2$ helpful \\
 \hline
IDA* ($LB_1$)           & 372 & 853,094,579 &  &  \\
IDA* ($LB_3$)           & 704 & 110,753,768 & &  \\
LIDA*                   & 368 & 130,695,270 & 42,862,888 & 19,060,111 \\
RLIDA*, $p_h=0.3$    & 337 & 233,077,220 & 27,628,566 & 13,575,017 \\
RLIDA*, $p_h\leq 0.5$  & 320 & 158,362,305 & 33,693,072 & 16,460,400 \\
Clairvoyant                 & 194 & 130,695,270 & 19,060,111 & 19,060,111 \\
\hline
\end{tabular}
\caption{Container Relocation}
\label{tbl:Container Relocation}
\end{centering}
\end{table}

In the experiments, we used as instances the 49 hardest tests out of those that were
solved in less than 20 minutes with the $LB_1$ heuristic,
from  the CVS test suite described in \cite{Caserta:2011:ACM:2039168.2039177,DBLP:conf/ieaaie/JinLZ13},
retrived from  http://iwi.econ.uni-hamburg.de/IWIWeb/Default.aspx?tabId=1083\&tabindex=4.
The instances actually used had either 5 or 6 stacks, and from 6 to 10 tiers.
Results are shown in table \ref{tbl:Container Relocation}.
In this domain Rational Lazy A* shows some performance improvement even when $p_h$ was assumed constant ($P_h=0.3$). However,
in this problem the branching factor is almost constant, and equal to the number of stacks minus 1, during much of
the search. As a result, there is room for improvement by better estimating $p_h$. Indeed using the bounds
developed in Section \ref{sec:bounds} to
estimate $p_h$ dynamically achieves significant additional speedup, as
shown by the line RLIDA*, $p_h\leq 0.5$. Due to the fact that the runtimes of the heuristics have a large variance
and are hard to predict precisely, using Eq. \ref{eqn:rlida-criterion-general} did not achieve good results, so
the results reported in the table are actually for the simplified decision rule that computes $h_2$ only when $p_hb(n) \geq 1$,
as mentioned after Equation \ref{eq:simplified}.

\section{DISCUSSION}

\subsection{Related work}\label{sec:related}

Other elaborate schemes for deciding on heuristics at runtime appear in the research literature.
Domshlak st al.~\cite{domshlak-et-al:jair-2012} also noted that although theoretically taking the maximum of
admissible heuristics is best within the context of A*, the overhead may not be worth it. Instead, their idea is to
select which heuristic to compute at runtime.
Based on this idea, they formulated {\em selective max} (Sel-MAX) for A*, an online learning scheme which
chooses one heuristic to compute at each
state.
In principle, Sel-MAX could be adapted to run in IDA*.
However, the domains we used in experiments
had a heuristic $h_1$ which has negligible computation time, and should thus always be computed. Sel-Max is
aimed at cases where there is a need for selection, i.e., if the time for computing each heuristic is not
negligible.

Automatically selecting combinations of heuristics for A* and IDA* from a large set of available heuristics
was examined in \cite{DBLP:conf/ausai/FrancoBR13}. Selecting a combination of heuristics is in some sense orthogonal to the work presented in
this paper, as once such a selection is made, one might still further optimize the actual scheme for computing the selected heuristics.
The heuristics can be evaluated lazily, and rationally omitting some of them conditional on the results of previously computed heuristics
in the same node can also be done. Generalizing both methods, one could try to optimize a {\em policy} for computing heuristics at the nodes, rather than
just find the best combination, but how to do so is non-trivial. That is because the number of policies is
at least doubly exponential in the number of heuristics under consideration, whearas the number of combinations is ``only'' exponential in the number of heuristics.

A related line of research of performing meta reasoning for IDA*-like algorithms
is on choosing the threshold for the next iteration.
In basic IDA*, the next threshold is strictly defined as the smallest value
among nodes that were pruned. Learning and decision making techniques are applied to
choose a different threshold such that time is saved but optimality of the algorithm is still
maintained~\cite{SarkarCGS91,Reinefeld, WahS94}.
This issue is orthogonal to the problem addressed in this paper.
In fact, our method for trading off time spent on computing heuristics with time spent on
expanding additional nodes should be extensible to other IDA*-like algorithms.
As in some of these algorithms the f-limit is not the next f-cost, such an extension
should overcome one of the major stumbling blocks to
further applicability of our method stated in Section \ref{sec:empirical}.

In addition, the notion of {\em type system} was recently introduced to divide
the state space into different types~\cite{DBLP:journals/ai/KorfRE01,ZahaviFBH10,LelisZH13}.
This was done usually for predicting the number of nodes expanded.
Our work here can be seen as using a simple type system for deciding whether to evaluate the $h_2$ heuristic.

\subsection{Summary and future work}

Rational Lazy IDA* and its analysis can be seen as an instance
of the rational meta-reasoning framework~\cite{RussellWefald}. While this framework
is very general, it is extremely hard to apply in practice. Recent work
exists on meta-reasoning in DFS algorithms for CSP) \cite{DBLP:conf/ijcai/TolpinS11}
and in Monte-Carlo tree search~\cite{DBLP:conf/uai/HayRTS12}. This paper applies
these methods successfully to a variant of IDA*.

We discussed two schemes for decreasing the time spent on computing heuristics during search.
Lazy IDA* is very simple and a natural implementation of IDA* in the presence of 2 or more heuristics,
especially if one is dominant but more costly.
Rational Lazy IDA* allows additional cuts in the number of $h_2$ computations, at the expense
of being less informed and thereby generating more nodes. However, due to a rational tradeoff, this
allows for an additional speedup, and Rational Lazy IDA* achieves the best
overall performance in our domains.

Experimental results on several domains show the advantage of RLIDA*.
The non-realizable clairvoyant scheme discussed in Section \ref{sec:empirical}
serves as a bound of the potential gain from RLIDA*.
We note that the most important term in some of the domains is $p_h$, the probability that $h_2$ will indeed
cause a cutoff. In this paper we provided a rudimentary method to bound $p_h$ based on previous
samples.
Future work might find better ways to estimate $p_h$, hopefully
getting closer to the clairvoyant ideal.
One such direction can be to use any of the newly introduced type-systems, e.g.,
those that measure the correlation of a given heuristic between neighbors~\cite{ZahaviFBH10,LelisZH13}.

Another direction is to relax some of the meta-reasoning assumptions, especially those frequently violated in practice,
and develop appropriate decision rules. In particular, consider the assumption that $h_1$ does not prune any of the children.
Preliminary runs on the tile puzzles showed that this assumption is violated in about 40\% of the nodes,
which seems to be a significant violation. Despite this violation, RLIDA* achieved most of the potential gain, so even though
relaxing this assumption may further improve the runtime, the extra effort (and possible runtime overhead) may not be worth it.
However, for the container relocation problem, this assumption was violated in about 60\% of the nodes and there is
also a considerable gap between RLIDA* and clairvoyant, so for this domain relaxing the assumption may be worth the effort.

Although the techniques used in this paper may be applicable to other IDA*-like algorithms
(e.g., RBFS, or DFBnB) the assumptions used in this paper are rather delicate,
necessitating a different set of assumptions and thus different resulting meta-level decision schemes
for such algorithm, another interesting item for future work.



\bibliography{paper}

\begin{thebibliography}{10}

\bibitem{Caserta:2011:ACM:2039168.2039177}
Marco Caserta, Stefan Vo{$\beta$}, and Moshe Sniedovich, `Applying the corridor
  method to a blocks relocation problem', {\em OR Spectr.}, {\bf 33}(4),
  915--929, (October 2011).

\bibitem{ASTR85}
R.~Dechter and J.~Pearl, `Generalized best-first search strategies and the
  optimality of {A}*', {\em Journal of the ACM}, {\bf 32(3)},  505--536,
  (1985).

\bibitem{domshlak-et-al:jair-2012}
Carmel Domshlak, Erez Karpas, and Shaul Markovitch, `Online speedup learning
  for optimal planning', {\em JAIR}, {\bf 44},  709--755, (2012).

\bibitem{INCJUR}
A.~Felner, U.~Zahavi, R.~Holte, J.~Schaeffer, N.~Sturtevant, and Z.~Zhang,
  `Inconsistent heuristics in theory and practice', {\em Artificial
  Intelligence}, {\bf 175}(9-10),  1570--1603, (2011).

\bibitem{DBLP:conf/ausai/FrancoBR13}
Santiago Franco, Michael~W. Barley, and Patricia~J. Riddle, `A new efficient in
  situ sampling model for heuristic selection in optimal search', in {\em
  Australasian Conference on Artificial Intelligence}, eds., Stephen Cranefield
  and Abhaya~C. Nayak, volume 8272 of {\em Lecture Notes in Computer Science},
  pp. 178--189. Springer, (2013).

\bibitem{DBLP:conf/uai/HayRTS12}
Nicholas Hay, Stuart Russell, David Tolpin, and Solomon~Eyal Shimony,
  `Selecting computations: Theory and applications', in {\em UAI}, eds., Nando
  de~Freitas and Kevin~P. Murphy, pp. 346--355. AUAI Press, (2012).

\bibitem{DBLP:conf/ieaaie/JinLZ13}
Bo~Jin, Andrew Lim, and Wenbin Zhu, `A greedy look-ahead heuristic for the
  container relocation problem', in {\em IEA/AIE}, eds., Moonis Ali, Tibor
  Bosse, Koen~V. Hindriks, Mark Hoogendoorn, Catholijn~M. Jonker, and Jan
  Treur, volume 7906 of {\em Lecture Notes in Computer Science}, pp. 181--190.
  Springer, (2013).

\bibitem{BFID85}
R.~E. Korf, `Depth-first iterative-deepening: An optimal admissible tree
  search', {\em Artificial Intelligence}, {\bf 27}(1),  97--109, (1985).

\bibitem{DBLP:journals/ai/KorfRE01}
Richard~E. Korf, Michael Reid, and Stefan Edelkamp, `Time complexity of
  iterative-deepening-{A}$^{\mbox{*}}$', {\em Artif. Intell.}, {\bf 129}(1-2),
  199--218, (2001).

\bibitem{KorfT96}
Richard~E. Korf and Larry~A. Taylor, `Finding optimal solutions to the
  twenty-four puzzle', in {\em AAAI}, pp. 1202--1207, (1996).

\bibitem{LelisZH13}
Levi H.~S. Lelis, Sandra Zilles, and Robert~C. Holte, `Predicting the size of
  {IDA*}'s search tree', {\em Artif. Intell.}, {\bf 196},  53--76, (2013).

\bibitem{Reinefeld}
Alexander Reinefeld and Tony~A. Marsland, `Enhanced iterative-deepening
  search', {\em IEEE Trans. Pattern Anal. Mach. Intell.}, {\bf 16}(7),
  701--710, (July 1994).

\bibitem{RussellWefald}
Stuart Russell and Eric Wefald, `Principles of metereasoning', {\em Artificial
  Intelligence}, {\bf 49},  361--395, (1991).

\bibitem{SarkarCGS91}
Uttam~K. Sarkar, Partha~P. Chakrabarti, Sujoy Ghose, and S.~C.~De Sarkar,
  `Reducing reexpansions in iterative-deepening search by controlling cutoff
  bounds', {\em Artif. Intell.}, {\bf 50}(2),  207--221, (1991).

\bibitem{thayer:bss}
Jordan~T. Thayer and Wheeler Ruml, `Bounded suboptimal search: A direct
  approach using inadmissible estimates', in {\em Proceedings of the
  Twenty-second International Joint Conference on Artificial Intelligence
  (IJCAI-11)}, (2011).

\bibitem{TOLPIN2013}
D.~Tolpin, T.~Beja, S.~E. Shimony, A.~Felner, and E.~Karpas, `Toward rational
  deployment of multiple heuristics in a', in {\em IJCAI}, (2013).

\bibitem{DBLP:conf/ijcai/TolpinS11}
David Tolpin and Solomon~Eyal Shimony, `Rational deployment of {CSP}
  heuristics', in {\em IJCAI}, ed., Toby Walsh, pp. 680--686. IJCAI/AAAI,
  (2011).

\bibitem{WahS94}
Benjamin~W. Wah and Yi~Shang, `A comparative study of ida*-style searches', in
  {\em ICTAI}, pp. 290--296, (1994).

\bibitem{ZahaviFBH10}
Uzi Zahavi, Ariel Felner, Neil Burch, and Robert~C. Holte, `Predicting the
  performance of {IDA}* using conditional distributions', {\em J. Artif.
  Intell. Res. (JAIR)}, {\bf 37},  41--83, (2010).

\bibitem{Zhang:2010:IIA:1945758.1945763}
Huidong Zhang, Songshan Guo, Wenbin Zhu, Andrew Lim, and Brenda Cheang, `An
  investigation of {IDA*} algorithms for the container relocation problem', in
  {\em Proceedings of the 23rd International Conference on Industrial
  Engineering and Other Applications of Applied Intelligent Systems - Volume
  Part I}, IEA/AIE'10, pp. 31--40, Berlin, Heidelberg, (2010). Springer-Verlag.

\end{thebibliography}
\bibliographystyle{ecai2014}

\end{document}